\pdfoutput=1
\documentclass[11pt]{article}

\usepackage{authblk}

\usepackage{acl2023}

\usepackage{times}
\usepackage{latexsym}
\usepackage{graphicx}
\usepackage{amsmath}
\usepackage{array}
\usepackage{lineno}

\usepackage[T1]{fontenc}
\usepackage[utf8]{inputenc}
\usepackage{bbm}
\usepackage{listings}
\usepackage{microtype}

\newcommand{\fooAlter}{\hspace{0pt}\textcolor{blue}{$\bullet$}\hspace{5pt}}
\newcolumntype{D}{ >{\centering\arraybackslash} m{2cm} }
\newcolumntype{M}[1]{>{\centering\arraybackslash}m{#1}}

\renewcommand\thefootnote{\ifcase\value{footnote}\or*\or\dag\or\ddag\fi}

\title{Applying RLAIF for Code Generation with API-usage in Lightweight LLMs}

\author[1, \dag]{Sujan Dutta}
\author[2]{Sayantan Mahinder}
\author[2]{Raviteja Anantha}
\author[2]{Bortik Bandyopadhyay}
\affil[1]{Rochester Institute of Technology, \texttt{sd2516@rit.edu}}
\affil[2]{Apple, \texttt{\{smahinder, raviteja\_anantha, bbandyopadhyay\}@apple.com}}

\begin{document}
\maketitle

\footnotetext[0]{\dag~Work done as a part of an internship at Apple.}

\begin{abstract}

Reinforcement Learning from AI Feedback (RLAIF) has demonstrated significant potential across various domains, including mitigating harm in LLM outputs, enhancing text summarization, and mathematical reasoning. This paper introduces an RLAIF framework for improving the code generation abilities of lightweight (<1B parameters) LLMs. We specifically focus on code generation tasks that require writing appropriate API calls, which is challenging due to the well-known issue of hallucination in LLMs. Our framework extracts AI feedback from a larger LLM (e.g., GPT-3.5) through a specialized prompting strategy and uses this data to train a reward model towards better alignment from smaller LLMs. We run our experiments on the Gorilla dataset and meticulously assess the quality of the model-generated code across various metrics, including AST, ROUGE, and Code-BLEU, and develop a pipeline to compute its executability rate accurately. Our approach significantly enhances the fine-tuned LLM baseline's performance, achieving a 4.5\% improvement in executability rate. Notably, a smaller LLM model (780M parameters) trained with RLAIF surpasses a much larger fine-tuned baseline with 7B parameters, achieving a 1.0\% higher code executability rate. \end{abstract}

\section{Introduction}
LLMs have demonstrated unprecedented natural language understanding and generation capabilities in recent times \cite{brown2020language, chowdhery2022palm, gpt4, anil2023palm, jiang2023mistral, touvron2023llama}. Reinforcement Learning with Human Feedback (RLHF) is a key contributor to this success. RLHF is a fine-tuning approach that uses human feedback to train models by incorporating human evaluations into the reward signal. This method improves model performance on complex tasks by aligning the model's behavior with human preferences. However, this technique is expensive due to the requirement for high-quality human feedback. RLAIF \cite{bai2022constitutional, lee2023rlaif} has emerged as a promising alternative to replace human feedback with AI feedback, making the fine-tuning more scalable. Concurrently, there is growing research interest in teaching LLMs how to use external tools (APIs) \cite{schick2024toolformer, nakano2021webgpt, patil2023gorilla, qin2023toolllm, li2023api, zhuang2024toolqa, hao2024toolkengpt}. However, the focus on lightweight models (<1B parameters) is limited. In this work, we propose an RLAIF framework to enhance lightweight LLMs' capability to generate code and effectively integrate API calls. Following Patil \textit{et al.} \shortcite{patil2023gorilla}, we consider the task of generating Python codes that include suitable API calls given instructions across a wide array of applications. The authors published the Gorilla dataset and showed that fine-tuned \texttt{LLaMA-7B} \cite{touvron2023llama} on this dataset outperforms non-finetuned LLMs like \texttt{GPT-4} \cite{gpt4} in terms of understanding a natural language request and mapping it to API calls. Using our RLAIF framework, we fine-tune \texttt{GPT-2-large} (780M parameters), which not only demonstrates comparable API call correctness to \cite{patil2023gorilla} but also surpasses its code generation performance.

\paragraph{Code Generation.} Although extensively studied since the early days of AI research, code generation \cite{waldinger1969prow, budinsky1996automatic, svyatkovskiy2020intellicode, li2022competition} remains a challenging problem. In recent years, the community has explored ways to apply RL in training machine learning models for code generation tasks. For instance, \textit{Seq2SQL} \cite{zhong2017seq2sql} proposed a neural network trained through RL for generating SQL queries given a text description. During training, a generated query is executed against a database, and the result is utilized as the reward in the RL algorithm. Le \textit{et al.} \shortcite{le2022coderl} developed \textit{CodeRL}, a sequence-to-sequence language model fine-tuned through an actor-critic RL approach for program synthesis. The code-generator LM is treated as the actor during the training, and the critic model, which is trained to predict unit test results, provides the reward for a generated code. Another work \cite{shojaee2023execution} similar to the above, proposed using feedback from code execution and a ground truth target code to compute the reward. While these approaches may perform well on classical programming tasks (e.g., writing SQL queries, solving competitive/interview-level coding problems, etc.), they are inapplicable on Gorilla-like \cite{patil2023gorilla} code generation where the program is required to load and execute ML models using the correct API. The bottleneck comes from the fact that the above-mentioned techniques require execution of the generated code to either compute the reward directly or train the critic model, but running thousands of such programs is prohibitively expensive.

\paragraph{Reinforcement Learning with AI Feedback.} Bai \textit{et al.} \shortcite{bai2022constitutional} introduced the concept of Reinforcement Learning with AI Feedback (RLAIF), which combines preferences labeled by LLMs with human-labeled preferences to optimize for helpfulness and harmlessness. Since then, many studies have explored the usefulness of AI-generated feedback as an alternative to expensive human annotations in various tasks. For instance, Luo \textit{et al.} \shortcite{luo2023wizardmath} proposed \textit{WizardMath}, which enhances the mathematical reasoning abilities of \texttt{Llama-2} using AI feedback in the training process. In another work \cite{zhang2023huatuogpt}, researchers used real-world data along with RLAIF to improve LLMs as medical consultants. Prior research has also explored AI evaluation for improving factual correctness in LLM-generated medical summaries \cite{mishra2023synthetic}. Kwon \textit{et al.} \shortcite{kwon2023reward} explored the usefulness of LLMs in the reward design for RL agents in Ultimatum Game, matrix games, and the DealOrNoDeal negotiation task. Recently, Lee \textit{et al.} \shortcite{lee2023rlaif} demonstrated that RLAIF can achieve human-level performance in summarization and helpful and harmless text generation. However, the possibility of using RLAIF to improve the code generation and API usage ability in small models (<1B parameters) is under-explored. We demonstrate that even with a few model parameters, AI feedback significantly improves code generation quality over simple fine-tuning baselines. Moreover, we found RLAIF applied on smaller 780M parameter \texttt{GPT-2-large} model outperforms \texttt{LLaMA-7B} fine-tuned models, which has nine times more parameters.

\section{Dataset}

\begin{figure*}[htb]
\centering
\includegraphics[scale = 0.6]{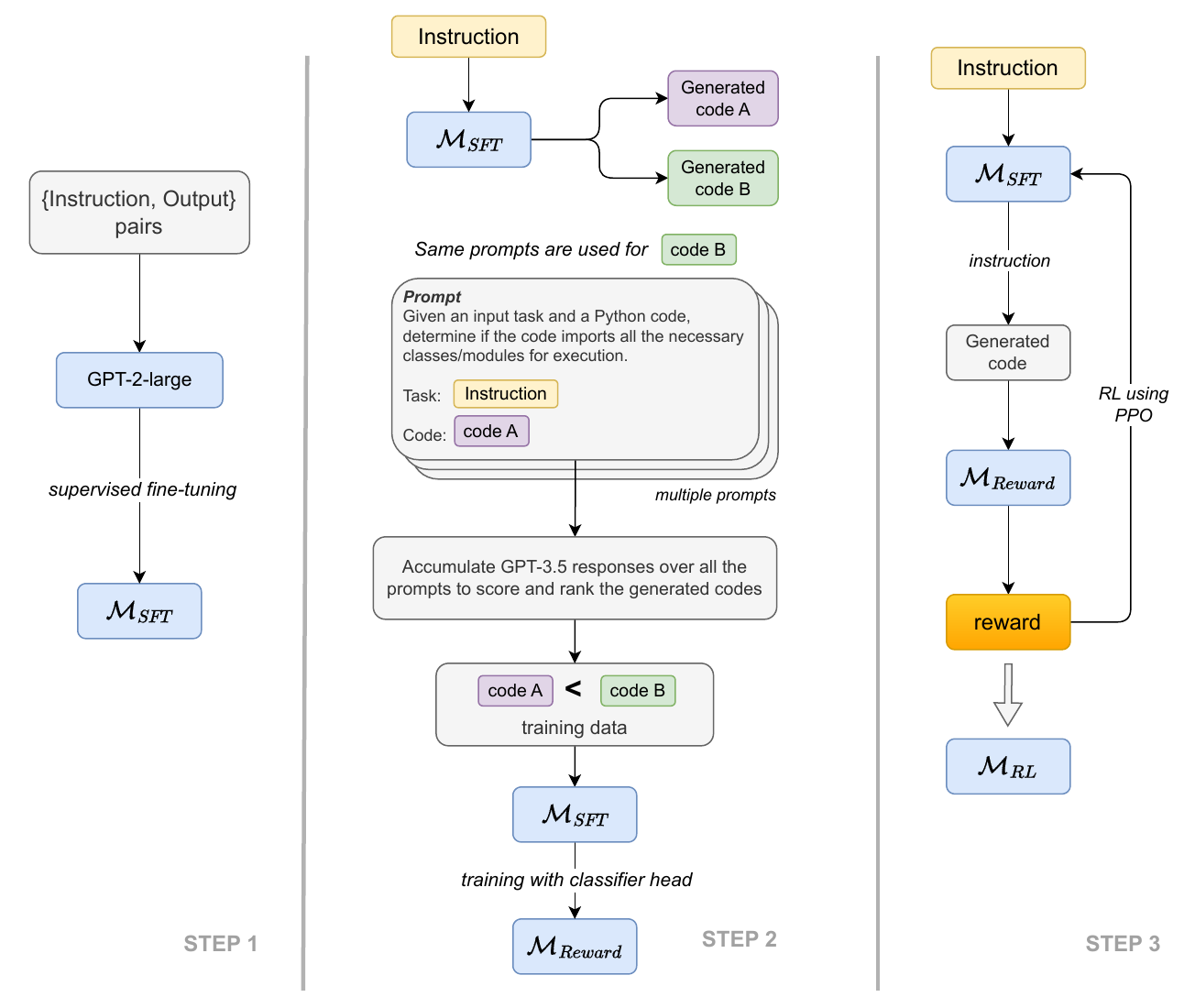}
\caption{Schematic diagram of the proposed framework. Step 1 is to fine-tune a base model on the dataset. In step 2, we score the $\mathcal{M}_\textit{SFT}$ generated outputs based on the \texttt{GPT-3.5} feedback using the technique described in section \ref{sec:method}. Using this score, we prepare preference data and train a reward model. Finally, in step 3, we use RL to fine-tune $\mathcal{M}_\textit{SFT}$ where $\mathcal{M}_\textit{reward}$ provides the reward.}
\label{fig:RLpipe}
\end{figure*}

We applied our proposed method to the Gorilla dataset published by Patil \textit{et al.} \shortcite{patil2023gorilla}. The Gorilla dataset consists of three parts - HuggingFace, TensorFlow, and PyTorch. In this work, we only focus on the HuggingFace dataset, which is the most extensive among the three, featuring over 925 unique APIs. These APIs belong to 37 different domains (e.g., Multimodal Text-to-Image, Computer Vision Image Classification, Audio Text-to-Speech, etc.), and for each API, there exist ten unique instructions. Each instance of the data contains an \emph{instruction} (task description), \emph{domain}, \emph{API call} (a single code line), \emph{explanation} (how to solve the task using the API), and a complete \emph{code} (Python script) to accomplish the task. Here, we highlight some key differences between Gorilla and the traditional code generation datasets. Most of the problem statements and corresponding code snippets present in the benchmark datasets including CodeSearchNet \cite{husain2019codesearchnet}, XLCoST \cite{zhu2022xlcost},  APPS \cite{hendrycksapps2021} and MBPP \cite{austin2021program} are related to traditional software engineering tasks, representative of common interview questions, require minimal computational resources to execute and do not require internet connection. On the contrary, the Python scripts in the Gorilla dataset focus on AI-related tasks and require an internet connection and significant computing resources (storage and processing power) to execute. The scripts are expected to download ML models hosted on HuggingFace, load them in memory, and run inference. So, techniques where code execution or unit test outcomes are treated as feedback \cite{zhong2017seq2sql,le2022coderl,shojaee2023execution} become inapplicable.

While Patil \textit{et al.} \shortcite{patil2023gorilla} focused only on generating the API call, we demonstrate the effectiveness of our approach both on API call correctness and the ability to use that API in a complete code. 

\section{Methodology}
\label{sec:method}

Our framework follows a similar pipeline to RLHF \cite{ouyang2022training}. However, instead of asking human annotators to rank the generated responses, we employ a bigger LLM by using a novel prompting strategy. More specifically, for a given instruction and generated code (containing an API call), we ask multiple binary (\textit{yes/no}) questions that capture different aspects of the generated code (and API call) to determine its quality. Our intuition is, that while generating code from natural language might be still challenging for LLMs, providing binary (\textit{yes/no}) answers guided by few-shot exemplars is a much easier task. These feedbacks in turn could be aggregated as a preference ground truth to train the reward model in the RLHF \cite{ouyang2022training} process. Thus our approach eliminates the need for expensive human annotation cost. We describe the proposed framework (Figure \ref{fig:RLpipe}) in detail.

\noindent \fooAlter \emph{\textbf{Step 1: Training a base model}}

The first step in the pipeline is to fine-tune a language model on the dataset to get a base model. We choose \texttt{GPT-2-large} and train it on the Gorilla dataset using the supervised fine-tuning technique for causal language models. We denote the fine-tuned model by $\mathcal{M}_{SFT}$.

\noindent \fooAlter \emph{\textbf{Step 2: Training a reward model using LLM feedback}}

Instead of human feedback from annotators, we employed a bigger LLM to generate the labels for the reward model.

We realized that human graders while judging the correctness of a response, consider different aspects of the generated output. Based on this intuition, we created multiple prompts ($P_i$) that ask different questions ($Q_i$) for the same input-output pair. More specifically, we created a set of 8 questions which We feed as prompts to a state-of-the-art language model (GPT-3.5) to get a binary response. 
% For the task in consideration (code generation with the correct API call),  the authors collectively created a set of 8 questions. 
Each of these questions addresses a different desired quality (free of bugs, correct imports, no undefined variables, correct syntax, etc.) of the output relevant to the task. Step 2 in Figure \ref{fig:RLpipe} presents a sample prompt made using one of the questions. The appendix contains the complete list of prompts. As the questions are binary (\textit{yes/no}) in nature, we simply count the number of \textit{yes} replies by $\mathcal{M}_\textit{GPT 3.5}$ to score each input-output pair. More formally, given a task $t$, generated output $o$, and question set $\{Q_i\}$ the prompt set is defined as $P(t, o) = \{P_i \mid P_i = [Q_i, t, o]\}$. The corresponding score ($S$) is given as: $$S(t, o) = \frac{\sum_{P_i \in P(t, o)}\mathbbm{I}(\mathcal{M}_\textit{GPT 3.5}(P_i)=\textit{yes})}{|P(t,o)|}$$ where $\mathbbm{I}$ is the indicator function and $\mathcal{M}_\textit{GPT 3.5}(P_i)$ is the reply from $\mathcal{M}_\textit{GPT 3.5}$ for the prompt $P_i$. We use this score to prepare the training data for $\mathcal{M}_\textit{reward}$ in the following way. For each instruction in the training data, we generate two outputs from  $\mathcal{M}_\textit{SFT}$ by varying the generation parameters (top-k, temperature, etc.). Then they are scored using the method described above and labeled (accept or reject) based on this score. These tuples of \{\emph{input instruction, accepted output, rejected output}\} are then combined to form the dataset for $\mathcal{M}_\textit{reward}$. In the training phase, $\mathcal{M}_\textit{reward}$ learns to classify whether a machine-generated code is acceptable (or not) for a given input instruction. We append a classifier head on top of $\mathcal{M}_\textit{SFT}$ and use this as the starting point of $\mathcal{M}_\textit{reward}$ and train for three epochs.

\noindent \fooAlter \emph{\textbf{Step 3: Reinforcement Learning}}

Finally, in the RL step, we fine-tune $\mathcal{M}_\textit{SFT}$ using the proximal policy optimization (PPO) algorithm \cite{schulman2017proximal}. The reward in this step is given by $\mathcal{M}_\textit{reward}$'s logit scores. We denote our final fine-tuned model by $\mathcal{M}_\textit{RL}$.

\section{Results and Discussions}

\begin{table}[!htb]
\begin{center}
    \resizebox{\columnwidth}{!}{\begin{tabular}{| p{3.2cm} | p{2cm} | p{1.5cm} | p{2cm} | p{1cm} |}
    \hline
  \centering {Model Name \\ (Size)}& \centering Executability Rate (\%) & \centering ROUGE $(\times 100)$ & \centering CodeBLEU $(\times 100)$ &  AST (\%) \\
    \hline
     \centering $\mathcal{M}_\textit{Gorilla}$ \shortcite{patil2023gorilla} (7B)&  \centering 26.9& \centering 41.2 & \centering 36.8 & 71.68 \\
    \hline 
    \centering $\mathcal{M}_\textit{SFT}$ (780M) & \centering 23.4& \centering 47.2 & \centering 40.6 &  {72.96}\\ 
    \hline
    \centering $\mathcal{M}_\textit{RL}$ (780M)&  \centering \textbf{27.9}& \centering \textbf{47.5} & \centering \textbf{42.2} & {\textbf{73.62}}\\
    \hline
    \end{tabular}}
    
\caption{Performance comparison of different models on the Gorilla dataset.}
\label{tab:results}
\end{center}
\end{table}

\begin{figure}[htb]
\centering
\includegraphics[scale = 0.5]{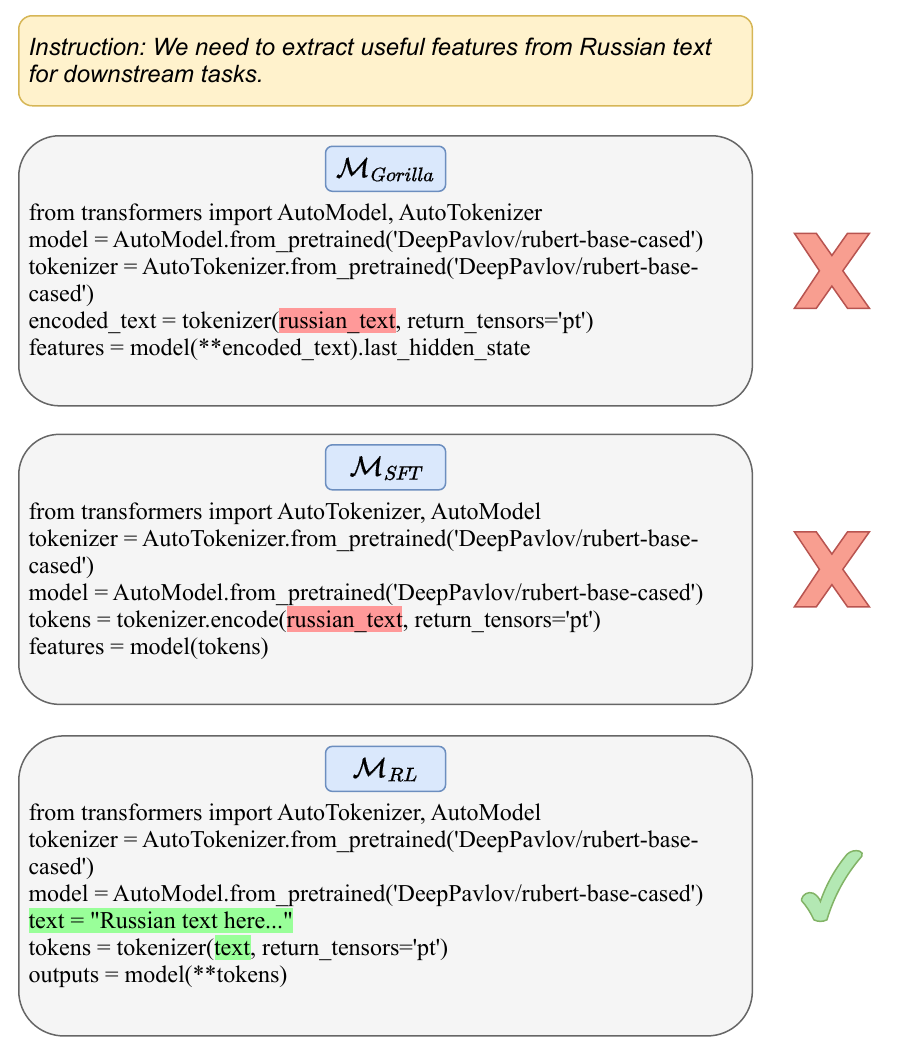}
\caption{Example code generated by different models for the same instruction. In the generations of $\mathcal{M}_\textit{Gorilla}$ and $\mathcal{M}_\textit{SFT}$ the variable \texttt{russian\_text} is undefined and hence will result in an error. Whereas $\mathcal{M}_\textit{RL}$ defines the variable \texttt{text} before using it.}
\label{fig:example}
\vspace{-.3cm}
\end{figure}

We compute the code generation quality using multiple metrics by comparing the generated output with the ground truth. The reported \emph{ROUGE} is the average of \emph{ROUGE-1, ROUGE-2, ROUGE-L, and ROUGE-sum} metrics introduced in \cite{lin2004rouge}. \emph{CodeBLEU} \cite{ren2020codebleu} was specifically designed for evaluating code synthesis. Ren \textit{et al.} \shortcite{ren2020codebleu} defined \emph{CodeBLEU} as the weighted average of standard \emph{BLEU} \cite{papineni2002bleu}, the weighted n-gram match ($\textit{BLEU}_\textit{weight}$), the syntactic AST match ($\textit{Match}_\textit{ast}$), and the semantic dataflow match ($\textit{Match}_\textit{df}$). $\textit{CodeBLEU} = \alpha \cdot \textit{BLEU} + \beta \cdot \textit{BLEU}_\textit{weight} + \gamma \cdot \textit{Match}_\textit{ast} + \delta \cdot \textit{Match}_\textit{df}$ . We set $\alpha=\beta=\gamma=\delta=0.25$ to give equal importance to all the components. The AST sub-tree-matching metric was proposed in \cite{patil2023gorilla} to capture the correctness of the API calls. In addition to that, we also report the successful execution rate of the generated code (\emph{Executability Rate}). It is worth noting that running this amount of machine-generated programs that download and use large AI models is challenging. We created a pipeline to automatically run the machine-generated codes in an isolated environment. 

Table \ref{tab:results} compares the performance of the proposed model with $\mathcal{M}_\textit{Gorilla}$ (finetuned \texttt{LLaMA-7B}) \shortcite{patil2023gorilla}. The results clearly show that the proposed $\mathcal{M}_\textit{RL}$ boosts the 
 performance of the supervised fine-tuned $\mathcal{M}_\textit{SFT}$ in terms of \emph{CodeBLEU} (1.6 points abs), \emph{AST} (0.66\% abs) and \emph{Executability Rate} (4.5\% abs). We also note that $\mathcal{M}_\textit{RL}$ outperforms the $\mathcal{M}_\textit{Gorilla}$ despite having only 1/9-th of the parameters. It is also reflected in the \emph{Executability Rate} of the generated code. Figure \ref{fig:example} shows an instance where our framework helps in fixing a common error present in $\mathcal{M}_\textit{Gorilla}$ and $\mathcal{M}_\textit{SFT}$ generations.

\section{Ethics statement}
This work adheres to the ethical guidelines and principles set out in the ACM Code of Ethics and followed by the broader research community. The dataset used in this paper was originally collected from public repositories hosted on HuggingFace. The authors are aware of the growing literature on jailbreaking language models to generate unsafe content. We hope the community will use the proposed models responsibly and only for the intended use cases.

\section{Limitations}
One of the common limitations faced by similar fine-tuned models is the presence of biases inherited from the pre-trained model. We anticipate that the biases present in the chosen base model (\texttt{GPT-2-large}) also exist in the final model $\mathcal{M}_\textit{RL}$ which might lead to the generation of biased code comments.

Another limitation of this work is the lack of diversity in programming language. The public dataset we considered contains only Python code. Future work should consider expanding this approach to encompass additional programming languages such as C++, Java, JavaScript, etc. Besides, we have not analyzed the performance between more frequent APIs (head) and infrequent APIs (tail). There might be some scope for improvements by focusing on tail APIs more. 

Lastly, the learning methodology applied in this study is offline. Given the rapid evolution and proliferation of machine learning models and the corresponding APIs for specific tasks, the model may not leverage more suitable APIs that emerge post-training. To address this, periodic updates to the model are necessary. Our framework's reliance on machine-generated feedback significantly reduces the resource intensity associated with the RLHF process, making these updates more feasible and less costly than a human feedback-based approach.

\bibliography{anthology,custom}
\bibliographystyle{acl_natbib}

\appendix
\section{Prompts}
Table \ref{tab:pset} lists all the prompts used in accessing the quality of the generated codes. 
\begin{table}[!htb]
\begin{center}
    \begin{tabular}{| p{8cm} |}
    \hline
    \textbf{Prompt}\\
    \hline
  \texttt{Given an input task and a Python code, determine if the code is functional.\newline
  TASK: [instruction] \newline
  CODE: [code]}\\
  \hline
  \texttt{Given an input task and a Python code, determine if the code imports all the necessary classes/modules for execution.\newline
  TASK: [instruction] \newline
  CODE: [code]}\\
  \hline
  \texttt{Given an input task and a Python code, determine if the code uses the correct functions/APIs.\newline
  TASK: [instruction] \newline
  CODE: [code]}\\
  \hline
  \texttt{Given an input task and a Python code, determine if the code is free of bugs and code smells.\newline
  TASK: [instruction] \newline
  CODE: [code]}\\
  \hline
  \texttt{Given an input task and a Python code, determine if the code is sufficient to accomplish the task.\newline
  TASK: [instruction] \newline
  CODE: [code]}\\
  \hline
  \texttt{Given an input task and a Python code, determine if the code uses indentations correctly.\newline
  TASK: [instruction] \newline
  CODE: [code]}\\
  \hline
  \texttt{Given an input task and a Python code, determine if the code uses quotes in string literals correctly.\newline
  TASK: [instruction] \newline
  CODE: [code]}\\
  \hline
  \texttt{Given an input task and a Python code, determine if the code uses duplicate parameters in a function.\newline
  TASK: [instruction] \newline
  CODE: [code]}\\
  \hline
\end{tabular}
    
\caption{Complete set of prompts. The tokens \texttt{[instruction]} and \texttt{[code]} are used to denote an instruction from the dataset and the corresponding generated code respectively.}
\label{tab:pset}
\end{center}
\end{table}

\section{Experimental details}
\subsection{Dataset}
The HuggingFace part of the Gorilla dataset \cite{patil2023gorilla} consists of over 9k instruction-output pairs. We trained our model on 90\% of the data and kept the rest for evaluation.

\subsection{Model and implementation details} 
$\mathcal{M}_\textit{SFT}$, $\mathcal{M}_\textit{reward}$ and $\mathcal{M}_\textit{RL}$ all have 780M parameters. While training $\mathcal{M}_\textit{SFT}$ and $\mathcal{M}_\textit{reward}$ we used a learning rate of $5 \times 10^{-4}$ and $5 \times 10^{-5}$ respectively. In the RL step (PPO algorithm), we set the learning rate to $6 \times 10^{-6}$. We did not perform any hyperparameter search. The results are reported by taking the mean of three inference runs. We implemented the training pipeline using the following Python libraries: transformers \cite{wolf2020transformers} and TRL \cite{vonwerra2022trl}.

\subsection{Computational cost}
We used a cluster of NVIDIA A100 40GB GPUs for our experiments. We spent in total $\sim$ 60 GPU hours for all of the experiments.
\end{document}